\documentclass[a4paper]{article}

\usepackage{INTERSPEECH2022}
\usepackage{multirow}

\title{Pseudo Label Is Better Than Human Label}
\name{Dongseong Hwang, Khe Chai Sim, Zhouyuan Huo, Trevor Strohman}
\address{Google, U.S.A}
\email{\{dongseong, khechai, zhhuo, strohman\}@google.com}

\begin{document}

\maketitle

\begin{abstract}
State-of-the-art automatic speech recognition (ASR) systems are trained with tens of thousands of hours of labeled speech data. Human transcription is expensive and time consuming. Factors such as the quality and consistency of the transcription can greatly affect the performance of the ASR models trained with these data. In this paper, we show that we can train a strong teacher model to produce high quality pseudo labels by utilizing recent self-supervised and semi-supervised learning techniques. Specifically, we use JUST (Joint Unsupervised/Supervised Training) and iterative noisy student teacher training to train a 600 million parameter bi-directional teacher model. This model achieved 4.0\% word error rate (WER) on a voice search task, 11.1\% relatively better than a baseline. We further show that by using this strong teacher model to generate high-quality pseudo labels for training, we can achieve 13.6\% relative WER reduction (5.9\% to 5.1\%) for a streaming model compared to using human labels.
% Human labeling is expensive. Labeling is the most painful step for ML production. It is widely believed that data is the new gold and big tech companies have an unfair advantage. Is it true that unlimited data unlimits model performance? In this study, we show $1$k hrs human labeled data is enough for the best ASR model. The model trained with $1$k hrs human labels and $26$k hrs pseudo labels has better Word error rate (WER) than the model with $27$k hrs human labels. Pseudo label training improves WERs of the production model by a significant margin; $5.9$ to $5.1$ on voice search. It means pseudo label quality is better than human label. To have quality pseudo labels, we utilized recent self/semi-supervised learning for a large ASR model.
\end{abstract}
\section{Introduction}

The success of end-to-end (E2E) speech recognition models \cite{sainath2020streaming,li2021better,narayanan2019recognizing} are highly dependent on having a large amount of high-quality transcribed speech data. However, it is expensive and difficult to obtain and maintain the high-quality human transcriptions, restricting the development of automatic speech recognition (ASR).  To relieve the dependence on large amounts of human-labeled data, self-supervised and semi-supervised learning techniques are heavily explored recently by leveraging on large-scale unlabeled speech data. 

Self-supervised learning techniques~\cite{schneider2019wav2vec,baevski2020wav2vec,chung2021w2v,chen2021injecting, hsu2021hubert,chiu2022self} have been shown to be effective for pre-training ASR models and achieved impressive performance for speech recognition tasks by leveraging large-scale unlabeled speech data. 
% \cite{chung2020generative,baevski2022data2vec}
In general, training is done in a 2-stage pretrain-finetune fashion where the models are first pre-trained using self-supervised learning using a large amount of unlabeled data and then fine-tuned using supervised data. For example, it was shown that Wav2vec 2.0~\cite{baevski2020wav2vec} pre-training can greatly reduce the amount of labeled data to just 100 hours and still achieve a reasonable word error rate (WER) performance of $4.0\%$ on the LibriSpeech test-other set~\cite{panayotov2015librispeech}. However, this is still worse than the model trained with all the training data. 
On the other hand, W2v-BERT~\cite{chung2021w2v} pre-training can achieve 2.8\% WER on the testother set with using the full 960 hours of training data.

%The results show that self-supervised pre-training can greatly reduce the requirement of the labeled data while still achieve without to achieve comparable performance as using the supervised training only. However, it is unclear how much labeled data we need. For example, W2v-BERT \cite{chung2021w2v} has LibriSpeech testother state-of-the-art (SoTA) WER ($2.8$) with full LibriSpeech data ($960$hrs). What is the lowest amount of labels to have the same SoTA WER? The WER with $100$ hrs labels by Wav2vec 2.0 \cite{baevski2020wav2vec} is $4.0$, which is far from the answer.

Another recent direction of research to leverage large-scale unlabeled data is semi-supervised learning, including noisy student training (NST) \cite{xie2020self,park2020improved} and consistency-based methods \cite{sohn2020fixmatch,xie2020unsupervised}. 
% \cite{berthelot2019mixmatch}
NST \cite{park2020improved} achieves the state-of-the-art (SoTA) performance on LibriSpeech when there are limited supervised speech data with transcriptions. YouTube model mostly relies on pseudo labels by NST~\cite{doutre2021improving}. To improve the performance further, \cite{zhang2021bigssl,hwang2021large} utilize a combination of self-supervised and semi-supervised learning as well as adding a third NST stage after the 2-stage pretrain-finetune scheme. Our previous paper~\cite{hwang2021large} demonstrates that the new approach reduces the requirement of human labels significantly, and matches or improves the SoTA performance with only $3\%$ human labels on the Google voice search task. However, it is still unclear whether a better performance can be achieved when more human labels are available. Otherwise, how much human labels should be mixed with pseudo labels to achieve the best performance?

In this paper, we will describe a multi-stage training strategy to train a strong teacher model to produce high-quality pseudo labels using a combination of self-supervised and semi-supervised learning, which has SoTA performance on voice search. We compare the effectiveness of human labels and pseudo labels by training the same model with corresponding transcriptions respectively. Our results show that the model trained with pseudo labels has significantly better WER of $5.1\%$ on voice search compared to that using human labels ($5.9\%$). This shows that the pseudo labels from a strong teacher have better quality than human labels for large-scale ASR model training. We also demonstrate that once we have a strong teacher model, subsequent training can be done entirely using only the pseudo labels.

The remainder of the paper is organized as follows. 
Section~\ref{sec:methods} describes the network architecture and the training methods used to train our strong teacher models for pseudo labeling. 
Section~\ref{sec:exp} shows the experimental results on the voice search and medium-form tasks.
Section~\ref{sec:discussion} presents some ablation studies and discussions.

%We also demonstrate that once we have a strong teacher, next teacher or production model needs only $1$k hrs human labels.

% The rest of the paper introduces model architecture and self-(semi-)supervised learning methods in Section  2,  describes  the experimental results in  Section  3, and discusses comparisons and analysis in Section  4.  We conclude in Section 5.

% \subsection{Related works}

% To solve the catastrophic forgetting and unstable training problem of the 2-stage pretrain-finetune scheme, Joint unsup and sup training (JUST) \cite{bai2021joint} co-trains the self-supervised and supervised losses together and achieves good performance. However, the mismatch between self- and supervised losses would degrade the learning of top layers with limited capacity, leading to performance regression on the supervised task.

% https://docs.google.com/drawings/d/1b9X-csqnxzKqZPj3hywK9UjTqBxjOFEv2noBcFbyMdc/edit
% https://docs.google.com/drawings/d/1_3-GjcZ2q37qEGSkS6XHm8nFOaE3DRWz5mUTI3Gv86A/edit

\section{Methods} \label{sec:methods}

\subsection{Model Architecture}
\label{sec:models}
We consider two types of models, one for the bi-directional teacher model and another for the streaming student model. 
Both models use the input feature vector of size $528$, consisting of $4$ contiguous frames of $128$-dimension log-mel features \cite{narayanan2019recognizing} sub-sampled by a factor of $3$ and a one-hot domain-id vector of size $16$. The outputs correspond to $4,096$ word pieces \cite{schuster2012japanese}.

The architecture and training procedure of the teacher model is similar to Conformer XL model ($0.6$B) in~\cite{zhang2020pushing}. The audio encoder has $24$ Conformer blocks~\cite{DBLP:conf/interspeech/GulatiQCPZYHWZW20} with model dimension $1024$. The convolution kernel size is $15$ and the self-attention layer consists of $8$ heads with $64$ left and right context length. The decoder consists of a 2-layer LSTM~\cite{hochreiter1997long} label encoder with $2048$ units projected down to $640$ output units, and a joint network with a single feed-forward layer with $640$ units. The total number of weights is $0.6$B. The model is trained by minimizing the RNN-T loss~\cite{Graves2012}. 

The student model uses cascaded encoders for unifying streaming and non-streaming ASR~\cite{narayanan2021cascaded}. The 1st-pass causal encoders has $7$ Conformer blocks with model dimension $512$. The 2nd-pass cascaded encoder has $10$ additional non-causal conformer layers that process a total of 900 milliseconds of future audio. Both causal and non-causal encoders feed into a shared hybrid autoregressive transducer (HAT) decoder~\cite{variani2020hybrid}. This model has a total of number $155$M weights.

We use a combination of self-supervised and semi-supervised learning to train a strong teacher model. Specifically, we combine the RNN-T~\cite{Graves2012} and W2v-BERT~\cite{chung2021w2v} losses using a hydra architecture to improve the teacher performance. We further improve the teacher model by applying multiple rounds of noisy student training~\cite{xie2020self}. We will describe these teacher training improvements in the following sub-sections. 

\subsection{JUST Hydra}
\label{sec:hydra}
\begin{figure}[t]
  \centering
  \includegraphics[width=\linewidth]{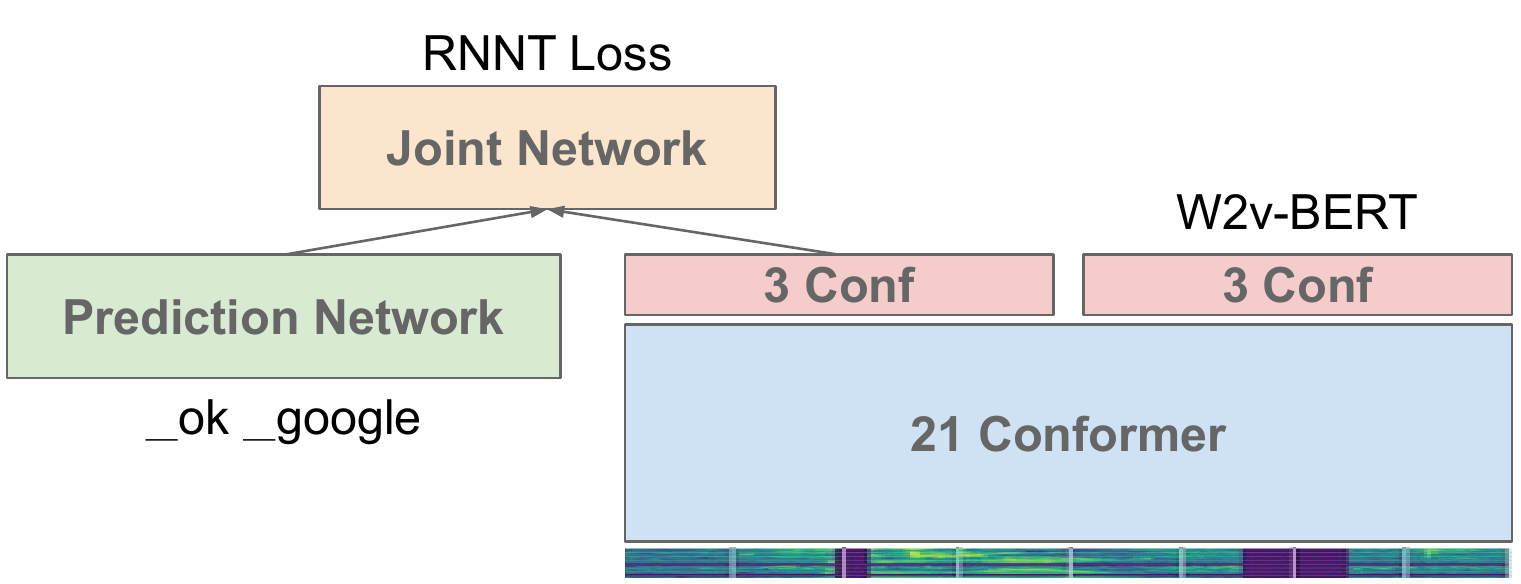}
  \caption{JUST Hydra -- a hydra architecture with separate layers for the last 3 Conformer layers for joint unsupervised and supervised training.}
  \label{fig:just_hydra}
\end{figure}

Previous work has shown that combining supervised and self-supervised learning in a multi-task learning fashion can be beneficial~\cite{hwang2021large,bai2021joint}. In this work, we use joint unsupervised/supervised training (JUST)~\cite{bai2021joint} to combine the supervised RNNT loss and the self-supervised W2v-BERT loss. JUST simplifies the 2-stage pretrain-finetune scheme into a 1-stage joint training. The representations from the last encoding layers are shared by both the RNNT joint network and W2v-BERT prediction layer. A recent study~\cite{pasad2021layer} shows that the last 3 layers in Wav2vec2.0~\cite{baevski2020wav2vec} are specialized to reconstructing masked inputs. This suggests that the last 3 layers should not be shared in JUST. Motivated by this work, we modify JUST to have separate layers for the for the last 3 Conformer layers, while most of lower layers are still shared, as shown in Fig.~\ref{fig:just_hydra}. 
We call it JUST hydra (JUST with Hydra heads).
JUST hydra uses both labeled and unlabeled data. The RNNT loss uses only the labeled data while the W2v-BERT self-supervised loss uses both the labeled and unlabeled data.

\subsection{Noisy student training (NST)}
In this study, we use noisy student training (NST)~\cite{xie2020self} in the same way as our previous study~\cite{hwang2021large}. 
When generating pseudo labels, we use a confidence estimation module (CEM)~\cite{qiu2021learning} to filter out low confidence utterances to avoid training with erroneous labels.
%When pseudo labels are generated, the teacher model filters out low confidence utterances by Confidence Estimation Module (CEM) , which gives the confidence score range $(0, 1)$. $1$ means $100\%$ confident for the utterance.
\begin{figure}[t]
  \centering
  \includegraphics[width=\linewidth]{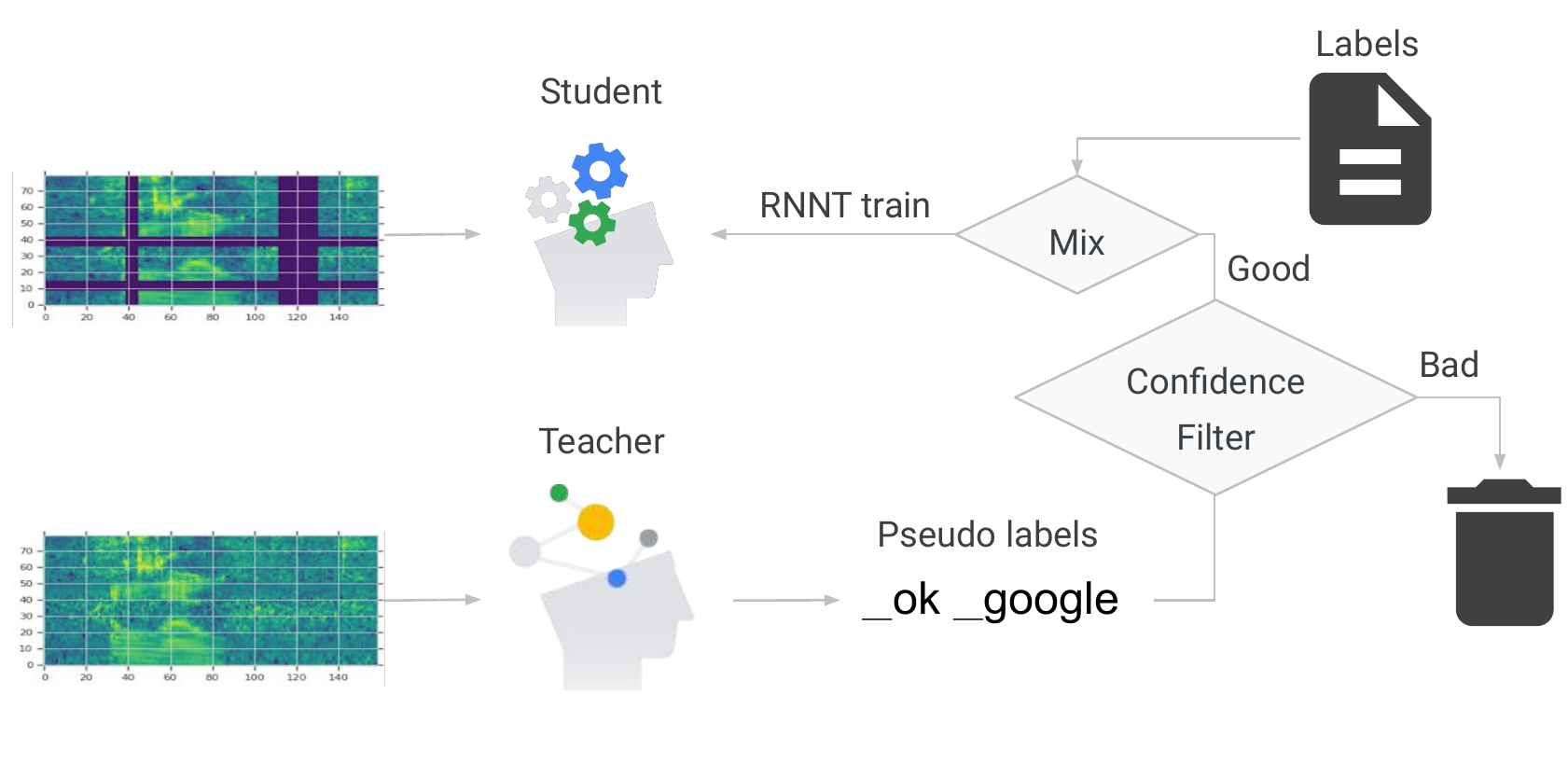}
  \caption{Noisy student training (NST).}
  \label{fig:nst}
\end{figure}
The filtered pseudo labels are then mixed with human-labeled data to form the training data to learn the student model using the RNNT loss, as shown in Fig.~\ref{fig:nst}.

We combine NST and JUST hydra as follows to progressively improve the teacher model:
\begin{enumerate}
\item Use JUST hydra to train a large ASR model using both the labeled and unlabeled data.
\item Using the above model as both the teacher and student, perform self-training to produce the 1st generation NST model.
\item Using the previous NST model as the teacher and initial JUST hydra model as the student, perform self-training to produce the 2nd generation NST.
\item Repeat NST until convergence.
\end{enumerate}

%Once we have the strong teacher model, we use NST to train the next generation teacher and production model. The next generation teacher is the teacher for out-of-domain (e.g. newer data). The production model is a smaller online ASR model for serving.
Once we obtain a strong teacher model, the student models are trained in a semi-supervised learning fashion using the pseudo labels generated by the teacher model.
\section{Experiments} \label{sec:exp}

\subsection{Data}
\label{sec:data}
We use a large multi-domain (\texttt{MD}) English dataset~\cite{Narayanan2018} for training, which consists of utterances from multiple domains, such as voice search (\texttt{VS}), medium-form (\texttt{MF}) and YouTube (\texttt{YT}). All the data are anonymized and hand-transcribed following Google AI principles~\cite{googleai}. \texttt{VS} is mostly voice command. \texttt{MF} is mostly natural conversation. The \texttt{YT} labels are generated from YouTube video transcription with confidence filter~\cite{liao2013large}.
This forms a total of $575$k hours of data. In addition, we also use $1.15$M hours of unsupervised data for voice search (\texttt{VS\textsubscript{unsup}}) and medium-form (\texttt{MF\textsubscript{unsup}}). 
We also used another set of unsupervised voice data (\texttt{VS\textsubscript{unsup-new}}) with 150k hours of data collected at a later time to study the generalization of our teacher models to new data.  
A summary of the training data is given in Table \ref{tab:data}. 

JUST hydra uses all the data. NST uses supervised data or all the data. The average utterance lengths for the \texttt{VS}, \texttt{MF} and \texttt{YT} domains are 4.6 seconds, 10.4 seconds and 5 seconds, respectively. 
For evaluation, we report the WER performance of the voice search (\texttt{VS}) and medium-form (\texttt{MF}) tasks. 
%In the final experiments, we use 3\% \texttt{VS} as labeled data, which is 800 hours. NST produces pseudo labels for rest of \texttt{VS}, \texttt{MF}, \texttt{en\_x} and \texttt{YT} data. 
%These data setup gives us the best \texttt{VS} WER.

\begin{table}[t]
    \centering
    \caption{Overview of training data sets. \texttt{MD} denotes all the supervised data (\texttt{VS} + \texttt{MF} + \texttt{en\_x} + \texttt{YT}, $575$k hrs). \texttt{MD\textsubscript{unsup}} denotes the unsupervised data (\texttt{VS\textsubscript{unsup}} + \texttt{MF\textsubscript{unsup}}, $1.15$M hrs). \texttt{en\_x} includes both \texttt{VS} and \texttt{MF}.}
    \label{tab:data}
    \begin{tabular}{ccc}
        \toprule
        \textbf{Data set} & \textbf{Label} & \textbf{Hours} \\
        \midrule
         Voice search (\texttt{VS}) & human & $27$k \\
         Medium-form (\texttt{MF}) & human & $26$k \\
         Dialect (\texttt{en\_x}) & human & $80$k \\
         Youtube (\texttt{YT}) & semi & $440$k \\
        \midrule
         Voice search (\texttt{VS\textsubscript{unsup}}) & pseudo & $900$k \\
         Medium-form (\texttt{MF\textsubscript{unsup}}) & pseudo & $250$k \\
         Voice search (\texttt{VS\textsubscript{unsup-new}}) & pseudo & $150$k \\
        \bottomrule
    \end{tabular}
    \vspace{-0.4cm}
\end{table}

\subsection{Results}

\subsubsection{Bi-directional teacher model}
\label{sec:teacher}
The teacher model is a $0.6$B-parameter bi-directional RNNT model. $B0$ in Table~\ref{tab:teacher} is the supervised baseline which is trained with \texttt{MD} using the RNNT loss~\cite{bagby2018efficient}. \texttt{E1} is trained using JUST Hydra, where the RNNT loss uses \texttt{MD}, and the W2v-BERT loss uses both \texttt{MD} and \texttt{MD\textsubscript{unsup}}. It improves the \texttt{VS} WER from $4.5$ to $4.3$.
On top of \texttt{E1}, we conducted multiple-generation NST to train \texttt{E2}, \texttt{E3} and \texttt{E4} using only the \texttt{MD} data with pseudo labels generated by a teacher model. \texttt{E2} is self-training, where both the teacher and student are based on \texttt{E1}. Subsequent training uses the previous iteration model as teacher and the student model is initialized with \texttt{E1}.
It is clear from Table~\ref{tab:teacher} that multiple-generation NST progressively improved the quality of the pseudo labels to yield better teacher models.
We obtained the best teacher model after two NST iterations, achieving 4.0 WER on \texttt{VS} and 2.0 on \texttt{MF}.
The best performance obtained using human labels is 4.3 and 2.3 on \texttt{VS} and \texttt{MF}, respectively.
This shows that using pseudo labels is better than using human labels. 
In the following section, we will use this teacher model to train production-style streaming models.

%It enhances \texttt{VS} WER from $4.3$ to $4.1$. We did the 2nd generation NST on \texttt{E2}, which is \texttt{E3}. The teacher is \texttt{E2} and the student is \texttt{E1}. It enhances \texttt{VS} WER further from $4.1$ to $4.0$. This is saturation point. We tried the 3rd generation NST, but the WERs are same as the 2nd generation.

%Multi-generation NST enhances \texttt{VS} WER from $4.3$ to $4.0$. The labels4.12.1E14student with pseudo labels by the old teacher becomes a better teacher, which produces better pseudo labels. \texttt{MD} is supervised data. The best WER by human labels is $4.3$ and the best WER by pseudo labels is $4.0$. Pseudo label is better than human label. \texttt{E3} has SoTA \texttt{VS} and \texttt{MF} WERs in Google. This will be used as teacher for the production model.

\begin{table}[t]
    \centering
    \caption{WERs of large models for supervised baseline, JUST hydra and NST.} 
    \label{tab:teacher}
    \begin{tabular}{lcccc}
        \toprule
        {\multirow{2}*{\textbf{Algorithms}}} & \multirow{2}*{\textbf{Data}} & \multicolumn{2}{c}{\textbf{WER ($\%$)}} \\
         & {} & {\textbf{VS}} & {\textbf{MF}} \\
        \midrule
        {B0 (Supervised)} & \texttt{MD} & ${4.5}$  & {${2.4}$} \\
        \midrule
        E1 (JUST hydra) & \texttt{MD} + \texttt{MD\textsubscript{unsup}} &  ${4.3}$ & {$2.3$} \\
        \midrule
        E2 (1st Gen NST, teacher=E1)  & {\texttt{MD}} & ${4.1}$  & {${2.1}$} \\
        E3 (2nd Gen NST, teacher=E2) & {\texttt{MD}} & $\textbf{4.0}$  & {$\textbf{2.0}$} \\
        E4 (3rd Gen NST, teacher=E3) & {\texttt{MD}} & $\textbf{4.0}$  & {$\textbf{2.0}$} \\
        \bottomrule
    \end{tabular}
    %\vspace{-0.4cm}
\end{table}

\subsubsection{Streaming student model}
\label{sec:prod}
The student model is a $155$M-parameter cascaded RNNT model. 
In Table \ref{tab:prod}, we compare two baseline models trained with \texttt{MD} data using human labels. 
The \texttt{B0\textsubscript{stream}} baseline trained with the RNNT loss achieved 5.9\% and 3.2\% WER on \texttt{VS} and \texttt{MD}, respectively.
Edit-based minimum Bayes risk (EMBR)~\cite{prabhavalkar2018minimum} further improved the performance to 5.5\% and 2.9\%.
When we perform NST on top of the \texttt{B0\textsubscript{stream}} model, using pseudo labels generated by the strong teacher, \texttt{E2}, from the previous section, we can improve the SoTA performance on \texttt{VS} and \texttt{MF} by a significant margin (5.1\% and 2.6\%).
Note that \texttt{B0\textsubscript{stream}} and \texttt{E1\textsubscript{stream}} have the same architecture and training hyper-parameters. The only difference is that the former is trained with human labels while the latter with pseudo labels. This shows that using pseudo labels is better than using human labels again. We conducted user A/B tests for \texttt{B0\textsubscript{stream}} and \texttt{E1\textsubscript{stream}}. \texttt{E1\textsubscript{stream}} won by 30:8 with ${<.1\%}$ p-value (${99.9\%}$ confidence).

EMBR training on top of \texttt{E1\textsubscript{stream}} (\texttt{E2\textsubscript{stream}}) did not yield further improvement. In fact, we observed degradation on \texttt{VS}. %It could be because EMBR computes the edit-distance based on human labels, which is worse than pseudo labels.

% Using NST enhances \texttt{VS} WER of the prod model by significant margin in \texttt{E1\textsubscript{stream}} from $5.9$ to $5.1$. It is the new prod SoTA WERs. It is quite surprising, because \texttt{B0\textsubscript{stream}} and \texttt{E1\textsubscript{stream}} are not very different but WERs are very different. Both models have the same architecture, use the same loss and data. Only difference is human labels vs pseudo labels. With pseudo labels, the model becomes significantly better. It means pseudo label quality is better than human label quality.

%We further use EMBR on \texttt{E1\textsubscript{stream}}, but it worsens WERs in \texttt{E2\textsubscript{stream}}. It is because EMBR computes edit-distance based on human labels, which is worse than pseudo labels.

\begin{table}[t]
    \centering
    \caption{WERs comparison for streaming models using supervised and NST training.} 
    \label{tab:prod}
    \begin{tabular}{lcc}
        \toprule
        {\multirow{2}*{\textbf{Algorithms}}}& \multicolumn{2}{c}{\textbf{WER}} \\
         & {\textbf{VS}} & {\textbf{MF}} \\
        \midrule
        {B0\textsubscript{stream} (Supervised)} & ${5.9}$  & {${3.2}$} \\
        {B1\textsubscript{stream} (B0\textsubscript{stream} + EMBR)} & ${5.5}$  & {${2.9}$} \\
        \midrule
        {E1\textsubscript{stream} (B0\textsubscript{stream} + NST)} & $\textbf{5.1}$  & {$\textbf{2.6}$} \\
        {E2\textsubscript{stream} (E1\textsubscript{stream} + EMBR)} & ${5.5}$  & {${2.6}$} \\
        \bottomrule
    \end{tabular}
    %\vspace{-0.4cm}
\end{table}

\section{Discussion} \label{sec:discussion}

\subsection{JUST hydra vs W2v-BERT}
\label{sec:just_hydra}

\begin{table}[t]
    \centering
    \caption{WERs for supervised baseline and W2v-BERT learning.} 
    \label{tab:selfsup}
    \begin{tabular}{lccc}
        \toprule
        \multirow{2}*{\textbf{Algorithms}}& \multirow{2}*{\textbf{Data}} & \multicolumn{2}{c}{\textbf{WER ($\%$)}} \\
         & {} & {\textbf{VS}} & {\textbf{MF}} \\
        \midrule
        {B0 (Supervised)} & \texttt{MD} & ${4.5}$  & {${2.4}$} \\
        \midrule
        E5 (W2v-BERT)  & \texttt{MD} + \texttt{MD\textsubscript{unsup}} & ${4.5}$  & {${2.5}$} \\
%        E6 (JUST)  & \texttt{MD} & ${4.7}$  & {${2.6}$} \\
        E6 (JUST)  & \texttt{MD} + \texttt{MD\textsubscript{unsup}} & ${4.5}$  & {${2.5}$} \\
        \midrule
        E1 (JUST hydra) & \texttt{MD} + \texttt{MD\textsubscript{unsup}} &  $\textbf{4.3}$ & {$\textbf{2.3}$} \\
        \bottomrule
    \end{tabular}
%    \vspace{-0.4cm}
\end{table}
W2v-BERT~\cite{chung2021w2v} for self-supervised learning has been shown to achieve state-of-the-art performance on LibriSpeech~\cite{panayotov2015librispeech}. However, we found that the vanilla W2v-BERT (\texttt{E5}) is worse than the baseline (\texttt{B0}), as shown in Table~\ref{tab:selfsup}. 
%We assume it is because of the domain-id issue as we discussed in \ref{sec:hydra}.
To improve the performance, we used JUST hydra, as described in Section~\ref{sec:hydra}. Instead of using W2v-BERT self-supervised learning for pre-training, we use JUST~\cite{bai2021joint} to combine RNNT and W2v-BERT losses in a multi-task learning setting.
However, using JUST alone (\texttt{E6}) did not improve the performance. With a hydra architecture, we use a separate 3-layer Conformer block to learn the representation for W2v-BERT. As a result, we were able to improve the WER on \texttt{VS} from $4.5$ to $4.3$ (\texttt{E1}).

As previously described in Section~\ref{sec:models}, the input log-mel features are augmented with a 16-dimensional one-hot domain embedding vector to better handle multiple domains. Since all the feature frames within an utterance are appended with the same domain embedding vector, it does not provide useful information for the W2v-BERT contrastive loss when the negative samples are extracted from the same utterance (used in this paper).
When the negative samples come from different utterances within a batch, the domain embedding is too easy to predict. In both cases, the domain information makes W2v-BERT training less effective. JUST hydra (\texttt{E1}) helps to alleviate this problem because jointly training with the RNNT loss can guide the model to learn meaningful domain embedding from data. The Hydra architecture is optimal multi-task architecture fully utilizing both W2v-BERT and RNNT losses.

%Our production ASR model is a multi-domain model to serve multiple services by one E2E model such as voice search, YouTube, Google meets, Google cloud, etc. We feed a domain-id \cite{sainath2020streaming} to the E2E model as a one-hot vector. The one-hot vector is concatenated with log-mel features. Domain-id is not useful information for W2v-BERT pretrain. When the negative sampling for the contrastive loss is from the same batch (which is the default), domain-id is redundant information as all frames have the same domain-id. When the negative sampling comes from different batch, domain-id is too obvious hint. In both cases, domain-id affects the self-sup badly.

%We use JUST \cite{bai2021joint} which is multi task learning including RNNT loss and W2v-BERT. As RNNT loss allows domain-id embedding to learn from data, we expect JUST has better WERs than the baseline. However, JUST (\texttt{E6}) does not enhance WERs. \texttt{E7} uses \texttt{MD} for RNNT loss and \texttt{MD} + \texttt{MD\textsubscript{unsup}} for W2v-BERT loss. When we use only \texttt{MD}, JUST (\texttt{E6}) is even worse than \texttt{B0}.
%That is why \ref{sec:hydra} introduces JUST hydra. JUST hydra finally fixes the domain-id issue. \texttt{E1} enhances \texttt{VS} WER from $4.5$ to $4.3$.

\subsection{Human label vs Pseudo label}
\label{sec:human_vs_pseudo}

\begin{table}[t]
    \centering
    \caption{WERs with different amounts of human labels and pseudo labels. All models are trained with \texttt{MD}.} 
    \label{tab:nst_xp}
    \begin{tabular}{lccc}
        \toprule
        \multirow{2}*{\textbf{Model}}& \multirow{2}*{\textbf{Human labels}} & \multicolumn{2}{c}{\textbf{WER ($\%$)}} \\
         & {} & {\textbf{VS}} & {\textbf{MF}} \\
        \midrule
        {E2 (Teacher)} & \multirow{2}*{All} & ${4.1}$  & {${2.1}$} \\
        {E1 (Student)} & & ${4.3}$  & {${2.3}$} \\
        \midrule
        E8 (NST)  & None & ${4.1}$  & {${2.1}$} \\
        E9 (NST)  & $330$ hrs & ${4.1}$  & {${2.1}$} \\
        E3 (NST, from Table~\ref{tab:teacher})  & $1$k hrs & $\textbf{4.0}$  & {$\textbf{2.0}$} \\
        E10 (NST)  & $2$k hrs & ${4.1}$  & {${2.1}$} \\
        E11 (NST)  & $4$k hrs & ${4.2}$  & {${2.1}$} \\
        E12 (NST)  & All & ${4.3}$  & {${2.3}$} \\
        \bottomrule
    \end{tabular}
    %\vspace{-0.4cm}
\end{table}
In Sections~\ref{sec:teacher} and \ref{sec:prod}, we have shown that using pseudo labels for training is better than using human labels.
Here, we investigate whether mixing a small fraction of human labels with pseudo labels could be beneficial.
In Table~\ref{tab:nst_xp}, we compared mixing $330$, $1$k, $2$k and $4$k hours of human labels when training the teacher model using NST. 
We observe that using $1$k hours of human labels achieve consistent improvements on both the \texttt{VS} and \texttt{MF} tasks.
WERs improved from 4.1\% to 4.0\% on \texttt{VS}, and from 2.1\% to 2.0\% on \texttt{MF}. The last mixing experiment (\texttt{E12}) uses only human labels, so WERs are same to the student (\texttt{E1}).

%We investigated the optimal amount of human-labeled data needed for NST. We found that $1$k hrs is sufficient for our task. JUST hydra provides competitive teacher (\texttt{E1}). On top of \texttt{E1}, NST further enhances \texttt{VS} WER to $4.0$ as \texttt{E3} shows in Table \ref{tab:teacher}. When we train NST (\texttt{E2}, \texttt{E3}), we actually use only $1$k hrs human labels, because it is the sweet spot. 

%We argued pseudo label quality is better than human label in \ref{sec:prod}. As we mix more human labels, the WERs are being regressed to the NST starting point (\texttt{E1}). Table \ref{tab:nst_xp} shows neither all human labels nor all pseudo labels are best. We should mix the certain amount of human labels, which is $1$k hrs. $1$k hrs human labels are enough. More human label hurts WERs. 

%we trained the production model with either $1$k hrs human labels or $100\%$ pseudo labels. Both gave the same prod SoTA WERs as shown in \texttt{E1\textsubscript{stream}} in Table \ref{tab:prod}.

On the other hands, mixing either none or $1$k hours labels with pseudo labels to train the streaming model achieved the same performance of 5.1\% and 2.6\%, as shown in Table \ref{tab:prod} (\texttt{E1\textsubscript{stream}}). Therefore, it is sufficient to rely 100\% on pseudo labels once we have a strong teacher model.
However, in practice, it may still be necessary to include a small amount of human labeled data. Completely relying on pseudo labels in multi-generation NST leads to deterioration of pseudo label quality as the teacher may reinforce confirmation bias in the pseudo labels.

%However, even though $100\%$ pseudo label model (\texttt{E8}) has good WERs, we still want to have the certain amount of human labels. %Otherwise, multi-generation NST totally relies on its own pseudo labels. Error is reinforced as the generation goes on. We want to have the certain amount of anchor points so that the model representation stays on real data manifold. We recommend to use $1$k hrs human labels to train new generation teacher on newer data. Distillation for the production model does not need human labels.

\subsection{Effect of initial student model}
We investigated how student models with different WER performance affect the final performance. 
In Table~\ref{tab:student_matter}, we compared NST training starting from three different initial student models.
\texttt{B1} is a poor starting model since it is trained with only \texttt{YT} (out-of-domain) data.
\texttt{B0} is trained using \texttt{MD} data with human labels. 
\texttt{E1} is trained with JUST hydra. 
In these experiments, we used 100\% pseudo labels for NST.
As shown in Table~\ref{tab:student_matter}, regardless of the initial student model, they all converged to the same performance of 4.1\% and 2.1\%.
We found that with a strong teacher model, NST is not sensitive the initial student model.
As shown in Section~\ref{sec:just_hydra}, it is important to apply self-supervised learning to train an initial teacher model.
Once we have a high-quality teacher, NST with pseudo labels is sufficient.

%Initial Student WER does not matter. Once we have a strong teacher, self-sup is not needed. Self-sup is one time bootstrap which produces the initial strong teacher. In \texttt{MD} data scale ($575$k hrs), self-sup does not provide anything more than (pseudo) labels for the downstream task.

%It is good news for the production engineering. Once we have the strong teacher, we do not need to use complicated self-sup training. RNNT training with non-user data (e.g. \texttt{YT}) is good enough for the initial student.

\begin{table}[t]
    \centering
    \caption{WERs of NST with $100\%$ pseudo labels based on various students.} 
    \label{tab:student_matter}
    \begin{tabular}{lccc}
        \toprule
        \multirow{2}*{\textbf{Model}}& \multirow{2}*{\textbf{Data}} & \multicolumn{2}{c}{\textbf{WER ($\%$)}} \\
         & {} & {\textbf{VS}} & {\textbf{MF}} \\
        \midrule
        {E2 (Teacher)} & & ${4.1}$  & {${2.1}$} \\
        \midrule
        {B1 (Supervised)} & \texttt{YT} & ${5.3}$  & {${4.0}$} \\
        {B0 (Supervised)} & \texttt{MD} & ${4.5}$  & {${2.4}$} \\
        E1 (JUST hydra) & \texttt{MD} + \texttt{MD\textsubscript{unsup}} &  ${4.3}$ & {$2.3$} \\
        \midrule
        E13 (NST, teacher=B1)  & \texttt{MD} w/o labels & ${\textbf{4.1}}$  & {${\textbf{2.1}}$} \\
        E14 (NST, teacher=B0)  & \texttt{MD} w/o labels & ${\textbf{4.1}}$  & {${\textbf{2.1}}$} \\
        E15 (NST, teacher=E1)  & \texttt{MD} w/o labels & ${\textbf{4.1}}$  & {${\textbf{2.1}}$} \\
        \bottomrule
    \end{tabular}
    %\vspace{-0.4cm}
\end{table}

\subsection{NST on out-of-domain data} \label{sec:ood}

So far, we have conducted NST experiments by replacing human labels with pseudo labels for the \texttt{MD} (\texttt{VS} + \texttt{MF} + \texttt{en\_x} + \texttt{YT}). We show the teacher to be able to generate better pseudo labels for the \texttt{MD} data. We investigated the pseudo label quality on out-of-domain (OOD) data by the teacher model trained on \texttt{MD}. We conducted ablation studies by performing NST on different OOD unsupervised data, as shown in Table~\ref{tab:ood}.
We considered two unsupervised voice search datasets (\texttt{VS\textsubscript{unsup}} and \texttt{VS\textsubscript{unsup-new}}).
The former has about 900k hours of data while the latter is a smaller (150k hours) but more recently collected data to capture temporal domain shift. Finally, we also created another \texttt{VS + MTR} dataset by artificially corrupting the clean utterances in \texttt{VS} using a room simulator, adding varying degrees of noise and reverberation with an average SNR of 12dB~\cite{kim2017generation}, a data augmentation technique used for multi-style training (MTR).

%We made the strong teacher (\texttt{E3}) using \texttt{MD} (\texttt{VS} + \texttt{MF} + \texttt{en\_x} + \texttt{YT}). How robust is this teacher for NST on domain shift? We did NST multiple students on out of domain (OOD) data in Table \ref{tab:ood}. Multi-style training (MTR) is augmented \texttt{VS} data. \texttt{VS\textsubscript{unsup}} is similar to \texttt{VS} but different data. \texttt{VS\textsubscript{new}} is recently collected data. \texttt{VS\textsubscript{new}} must have domain shift compared to \texttt{VS}. MTR data are created by artificially corrupting the clean utterances using a room simulator, adding varying degrees of noise and reverberation with an average SNR of 12dB~\cite{kim2017generation}.

We use NST to train both the large $0.6$B-parameter models (\texttt{Large}) and $155$M-parameter streaming models (\texttt{Stream}) with $100\%$ pseudo labels. From Table~\ref{tab:ood}, we observe that NST training for the \texttt{Large} model is fairly robust to the unsupervised data used. The \texttt{E3} teacher model (trained on \texttt{VS}) works well for other OOD data (\texttt{VS\textsubscript{unsup}}, \texttt{VS\textsubscript{unsup-new}}, \texttt{VS + MTR}) and achieve similar performance of 4.1--4.2\%. 
On the other hand, the \texttt{Stream} models performed slightly worse on OOD unsupervised data (5.4--5.5\%).
% the quality of the pseudo labels is slightly worse and 
This means that the \texttt{Stream} models are sensitive to domain shift. Large bi-directional model is better for unseen domain generalization.
Nevertheless, the OOD streaming models are still better than B0\textsubscript{stream}, the baseline model trained with human labels (5.9\%).

%It enables online learning. We can train incoming data by NST. Time to time, we replace the old teacher with the newer generation teacher.

%The large model keeps \texttt{VS} WER to $4.1$ with various domains, but the production model is regressed as shown in Table \ref{tab:ood}. A small model is sensitive for domain shift. It is why teacher model must be a large scale model.

%The amount of \texttt{VS} ($27$k hrs) is good enough for pseudo labeling. \texttt{VS} + \texttt{VS\textsubscript{unsup}} ($927$k hrs) does not give extra performance enhancement. It is good news for the engineering. We do not need to keep data as many as possible. There are the certain amount of data enough. It is different opinion compared to the LM scaling law; larger model and more data make performance better in Neural Language models \cite{kaplan2020scaling}. Discriminative models (e.g. ASR) may have diminishing return after the certain amount of data, unlike generative models (e.g. LM). We leave scaling ASR model and unsupervised data for future research.

\begin{table}[t]
    \centering
    \caption{WERs of NST with various data for large and production models.} 
    \label{tab:ood}
    \begin{tabular}{lcc}
        \toprule
        \multirow{2}*{\textbf{Data}}& \multicolumn{2}{c}{\textbf{VS WER ($\%$)}} \\
         & {\textbf{Large}} & {\textbf{Stream}} \\
        \midrule
        {E3 (Teacher)} & \multicolumn{2}{c}{${4.0}$} \\
        {B0, B0\textsubscript{stream} (Student)} & ${4.5}$  & ${5.9}$ \\
        \midrule
        \texttt{VS} & ${\textbf{4.1}}$  & {${\textbf{5.1}}$} \\
        \texttt{VS\textsubscript{unsup}} &  ${\textbf{4.1}}$ & {$5.4$} \\
        \texttt{VS\textsubscript{unsup-new}} &  ${4.2}$ & {$5.5$} \\
        \texttt{VS} + \texttt{MTR} & ${\textbf{4.1}}$  & {${5.5}$} \\
%        \texttt{VS} + \texttt{VS\textsubscript{unsup}} &  ${\textbf{4.1}}$ & {$5.3$} \\
        \bottomrule
    \end{tabular}
    \vspace{-0.4cm}
\end{table}

%\subsection{Pseudo label is better}
%You may ask why do not we improve human label quality. Google has made lots of efforts for human label quality, since Google voice search was launched in 2011. We have rigorous transcription policies to make human label consistent throughout worker and time. It comes out pseudo label is better even after all the efforts. Human error is inevitable. More investment is not practical for not only Google but also other companies.

%A teacher model is a statistical model. The pseudo label is the highest likelihood estimation (by beam search). It is more consistent than human. It is good news. Labeling is boring. It should be machine's job. All human should do is $1$k hrs labels.

\subsection{Limitations and future work}
This study demonstrates that a strong teacher can produce high-quality pseudo labels for NST. Once we have a strong teacher, we need only $1$k hrs human labels to produce the next generation best teacher model. However, human labels are still needed to train a good initial teacher model, as JUST hydra uses all human labels (\texttt{MD}) in Table \ref{tab:selfsup}. For languages that do not have sufficient human labeled data, it remains a challenge and open problem to train high quality initial teacher models using as little data as possible.

%can produce another strong teacher by NST with $1$k hrs human labels. Once we have a strong teacher, $1$k hrs human labels are enough to keep. However, JUST hydra used full human labels (\texttt{MD}) to produce the first teacher in \ref{sec:teacher}. In Table \ref{tab:w2b_xp} in \ref{sec:w2b_xp}, W2v-BERT with $1$k hrs human labels has far worse \texttt{VS} WER than JUST hydra with \texttt{MD} ($4.9$ vs $4.3$). The current SoTA self-supervised learning requires more human labels than $1$k hrs. Self-supervised learning is hot topic in speech ML communities and has been great progress. We leave self-sup learning with $1$k hrs for future research.

\section{Conclusions}
In this paper, we show that by using joint unsupervised and supervised training with W2v-BERT and noisy student training, we can train a strong teacher model that achieved 11.1\% relative word error rate improvement compared to a baseline model trained with human labeled data only.
This teacher can generate high-quality pseudo labels that are more consistent than the human labels. Semi-supervised learning with the pseudo labels can achieve 13.6\% relative word error rate improvement, compared to our best streaming model for the voice search task.

%We shows the recipe to produce a strong teacher model using W2b-BERT, JUST hydra and NST. We demonstrated pseudo labels from the strong teacher are better than human labels. We answered the million dollar question "how many human labels do we need?". The answer is $1$k hrs human labels.

\section{Acknowledgements}
We thank Françoise Beaufays, Zhehuai Chen, Chung-Cheng Chiu, Shefali Garg, Mohammadreza Ghodsi, Bo Li, Pedro Moreno Mengibar, Ananya Misra, Bhuvana Ramabhadran, Andrew Rosenberg, David Qiu, Tara Sainath, Hasim Sak, Nikhil Siddhartha, Yonghui Wu, Yu Zhang for helpful discussions.

% \newpage 
\bibliographystyle{IEEEtran}
\bibliography{main}

\newpage 
\appendix
\section{Appendix}

\subsection{NST based on supervised teacher}
When we start NST based on poor supervised baseline (\texttt{B0}), we couldn't make the SoTA teacher (\texttt{E3} in Table \ref{tab:teacher}) by multi generation NST. Self-sup is needed only once.

\begin{table}[ht]
    \centering
    \caption{WERs of teacher models for supervised baseline, JUST hydra and NST.} 
    \label{tab:nst_sup}
    \begin{tabular}{cccc}
        \toprule
        {\textbf{Algorithms}}& {\textbf{Data}} & \multicolumn{2}{c}{\textbf{WER ($\%$)}} \\
         & {} & {\textbf{VS}} & {\textbf{MF}} \\
        \midrule
        {B0 (Supervised)} & Sup & ${4.5}$  & {${2.4}$} \\
        \midrule
        E16 (B0, 1st Gen NST)  & Sup & ${4.4}$  & {${2.3}$} \\
        E17 (E16, 2nd Gen NST) & Sup & ${4.3}$  & {${2.3}$} \\
        E18 (E17, 3rd Gen NST) & Sup & ${4.3}$  & {${2.2}$} \\
        \bottomrule
    \end{tabular}
    \vspace{-0.4cm}
\end{table}

\subsection{W2v-BERT with the fraction of data}
\label{sec:w2b_xp}
In Table \ref{tab:w2b_xp}, we train both supervised baseline and W2v-BERT with none, $1$k hrs and full labeled data. 
%All WERs in W2v-BERT are not optimal due to the domain-id issue. 
In low resource regime (none and $1$k hrs labels), W2v-BERT enhances WERs compared to the baselines respectively. However, in high resource regime ($27$k hrs full labels), W2v-BERT contributes nothing.

Self-sup contribution has diminishing returns as the amount of labels is growing. The self-sup scale issue was found in computer vision domain as well \cite{ghiasi2021multi}. Self-sup outperforms to the supervised baseline with ImageNet (1.3M images), but self-sup has worse performance than the supervised baseline with JFT (300M images).

\begin{table}[ht]
    \centering
    \caption{WERs of W2v-BERT with the fraction of data.} 
    \label{tab:w2b_xp}
    \begin{tabular}{cccc}
        \toprule
        {\textbf{Algorithms}}& {\textbf{Amount of}} & \multicolumn{2}{c}{\textbf{WER ($\%$)}} \\
         & {\textbf{VS/MF Data}} & {\textbf{VS}} & {\textbf{MF}} \\
        \midrule
        \multirow{3}*{Supervised} & None & ${5.5}$ & ${4.8}$ \\
        {} & $1$k hrs & ${5.2}$ & ${2.9}$ \\
        {} & All & ${4.5}$ & ${2.4}$ \\
        \midrule
        \multirow{3}*{W2v-BERT} & None & ${5.2}$ & ${4.0}$ \\
        {} & $1$k hrs & ${4.9}$ & ${2.7}$ \\
        & All & ${4.5}$ & ${2.5}$ \\
        \bottomrule
    \end{tabular}
%    \vspace{-0.4cm}
\end{table}

\subsection{Confidence filter}
Our previous study \cite{hwang2021large} reported NST with confidence filter enhances WER further, because it filters out not-confident utterances whose pseudo label maybe wrong. It was true for small teacher ($120$M). However, confidence filter on the large teacher doesn't improve WERs. We searched the confidence threshold in range (1e-6, 1e-4, 1e-2, 0.1, 1, 5, 9, 9.9). None of thresholds gave better WERs. It actually often degraded the performance. Strong teacher doesn't need confidence filter.

\subsection{Hypothesis comparison}
In Section~\ref{sec:human_vs_pseudo}, the model trained with pseudo labels has better WERs. In Table \ref{tab:hyps}, we compare hypothesises from the models trained with human labels and pseudo labels respectively. The pseudo label model produces hypothesises respecting the transcription policy of Google, persistently. On the other hands, the human label model sometimes produces hypothesises which does not comply our normalization policy. It is because our human labels includes this kind of incorrect labels, while pseudo labels are more consistent as those are produced by a statistical model, which provides the most likely labels consistently. The second group in Table \ref{tab:hyps} shows the pseudo label model is generally better as well. We assume the human label model wastes model capacity by inconsistent normalization, so the general performance of the model is worse than the pseudo label model.

\begin{table}[ht]
    \centering
    \caption{Hypothesis comparison between models trained with human labels and pseudo labels.} 
    \label{tab:hyps}
    \begin{tabular}{cc}
        \toprule
        {\textbf{Model with human label}}& {\textbf{Model with pseudo label}} \\
        \midrule
        {OOO km per hour} & {OOO km/h} \\
        {OOO 8 and 1/2} & {OOO 8 1/2} \\
        {9 + 6/7 + 7 + 5/12} & {9 6/7 + 7 5/12} \\
        {OOO AC DC} & {OOO AC/DC} \\
        \midrule
        {OOO under water?} & {OOO underwater?} \\
        {OOO a County?} & {OOO accounting?} \\
        \bottomrule
    \end{tabular}
    \vspace{-0.4cm}
\end{table}

\end{document}